\documentclass[11pt]{article}

\usepackage[preprint]{acl}
\usepackage{times}
\usepackage{latexsym}
\usepackage[T1]{fontenc}
\usepackage[utf8]{inputenc}
\usepackage{microtype}
\usepackage{graphicx}
\usepackage{booktabs}

\title{Does the Truthfulness Signal Survive Code-Mixing? Probing Hidden States \\ for Hallucination Detection in Hinglish}

\author{
  Tanveer Singh \\
  Plaksha University \\
  \texttt{tanveer.singh.ug25@plaksha.edu.in} \\
}

\begin{document}
\maketitle

\begin{abstract}
Hidden-state hallucination probing - training a linear classifier on an LLM's
internal activations to detect whether a generated answer is faithful to the
input - is an active area of 2026 research, with recent work reporting
0.90--1.00 AUROC across several benchmarks and languages. However, none of
this work has tested probes on code-mixed input, despite the fact that a huge
population of chatbot users write in Hindi--English code-mixed text
(``Hinglish''). We address this gap directly: does a hallucination probe
trained on clean-language hidden states transfer to Hinglish, or does the
signal degrade under code-mixing? We construct a 5{,}674-item Hindi/English/
Hinglish QA benchmark, generate and label 17{,}022 model responses across
three open-weight 7--8B LLMs (Qwen2.5-7B, Mistral-7B, Llama-3.1-8B), extract
per-layer hidden states at two token positions, and train linear and MLP
probes for in-distribution detection and cross-lingual transfer. We find that
the hallucination signal survives code-mixing well: transfer AUROC ranges
from 0.88 to 0.99, with gaps of mostly under 0.05 AUROC relative to
in-distribution performance, and that Hindi-trained probes transfer to
Hinglish more reliably than English-trained probes. As an independent,
practically motivated finding, all three models hallucinate substantially
more on Hindi and Hinglish than on English for matched facts. We release our
code and synthetic Hinglish QA dataset to support further work on code-mixed
hallucination detection.
\end{abstract}

\section{Introduction}
\label{sec:intro}

Large language models (LLMs) are increasingly deployed as conversational
assistants for users who do not write in clean, monolingual English. A
particularly large such population types in \emph{Hinglish} - Hindi and
English mixed within the same sentence or utterance - when interacting with
chatbots, search assistants, and customer-service agents across South Asia.
Despite this, essentially all published work on detecting when an LLM is
hallucinating by probing its internal hidden states has been evaluated on
clean, single-language text.

Hidden-state probing for hallucination detection - training a lightweight
classifier (typically linear) on an LLM's per-layer activations to predict
whether a generated answer is faithful or hallucinated - has emerged as one
of the most effective and lightweight detection strategies available. Four
recent studies illustrate the strength of this approach: \citet{zhang2025icrprobe}
track how each layer contributes to the residual stream update rather than
probing static activations, and use this signal to detect hallucinations
with substantially fewer parameters than prior probes; \citet{mrykhin2026pep}
augment standard linear probes with a small number of learnable prompt
embeddings and show consistent gains over the vanilla-linear baseline, along
with promising pre-generation and cross-model transfer; \citet{alvi2026multihaludet}
extend hidden-state probing to a multilingual setting for the first time,
reporting robust cross-lingual generalization of a stacked probing framework
across French, Bangla, and Amharic; and \citet{aiersilan2026quantized} show
that even under aggressive 4-bit quantization, a single linear probe on a
mid-network layer recovers 0.904--1.000 AUROC across three 7--8B open-weight
model families, with MLP probes rarely improving on the linear baseline by
more than 0.01 AUROC. Taken together, this body of work establishes that
truthfulness is linearly decodable from hidden states, that the signal is
robust to quantization, and - per \citet{alvi2026multihaludet} - that it
can transfer across typologically distinct \emph{languages}. What remains
untested is whether the signal survives \emph{within-sentence} code-mixing,
where the model must track truthfulness across a single input that switches
languages at the word or phrase level, a qualitatively different and
arguably harder generalization test than transfer between monolingual
languages.

This gap is not a narrow technical oversight. Code-mixed input is the
default register of everyday typed communication for hundreds of millions of
Hindi--English bilingual speakers, and any hallucination detector intended
for real-world deployment in this population will encounter Hinglish
constantly. If the truthfulness signal that hidden-state probes rely on is
disrupted by code-mixing - for instance because code-mixed tokens sit
outside the distribution the base LLM was most heavily trained on, or because
switching languages mid-sentence perturbs the residual stream in ways that
are entangled with (rather than orthogonal to) the truthfulness direction -
then probes validated only on clean text could fail silently exactly where
they are needed most.

We study two research questions:
\begin{enumerate}
    \item[\textbf{RQ1}] (Primary) Does a hallucination probe trained on
    clean-language hidden states transfer to Hinglish, or does the
    truthfulness signal degrade under code-mixing?
    \item[\textbf{RQ2}] (Secondary) Do models hallucinate more on Hinglish
    than on clean Hindi or English for the same underlying facts?
\end{enumerate}

To answer these questions, we construct a QA benchmark spanning clean Hindi,
clean English, and synthetic Hindi--English code-mixed (Hinglish) items
derived from the same underlying facts, generate answers from three
open-weight 7--8B instruction-tuned LLMs (Qwen2.5-7B, Mistral-7B, and
Llama-3.1-8B) in all three language conditions, label each answer as
faithful or hallucinated, extract per-layer hidden states at two token
positions, and train linear and MLP probes for both in-distribution
detection and train-on-clean/test-on-Hinglish transfer.

We find that the hallucination signal survives code-mixing well: transfer
AUROC ranges from 0.88 to 0.99 across models and probing positions, with the
gap to in-distribution performance under 0.05 AUROC in most conditions, and
that probes trained on clean Hindi transfer to Hinglish more reliably than
probes trained on clean English. As a secondary, independently motivating
finding, all three models are substantially more likely to hallucinate when
answering in Hindi or Hinglish than in English on matched questions.

Our contributions are:
\begin{itemize}
    \item The first evaluation of hidden-state hallucination probing under
    within-sentence code-mixed input, addressing a gap left open by all
    four prior probing studies we survey.
    \item Evidence that the truthfulness signal transfers to Hinglish with
    only modest degradation, and that this holds across three model
    families and two probing positions.
    \item Evidence that Hindi-trained probes transfer to Hinglish better
    than English-trained probes, which we discuss as consistent with a
    shared-script/shared-vocabulary hypothesis.
    \item Release of our code and synthetic Hinglish QA dataset to support
    future work validating these findings on naturally-occurring code-mixed
    corpora.
\end{itemize}

\section{Related Work}
\label{sec:related}

\paragraph{Hidden-state hallucination detection.}
A growing line of work shows that an LLM's internal activations encode a
recoverable signal about whether its own output is truthful.
\citet{azaria2023internal} were among the first to demonstrate that a simple
classifier trained on hidden states can predict the truthfulness of a
statement more reliably than the model's own stated confidence.
\citet{orgad2024llms} extend this analysis, showing that truthfulness
information is concentrated in specific token positions rather than spread
uniformly across a sequence, and that probes trained on one dataset transfer
poorly to datasets requiring different underlying skills - a generalization
concern closely related to, but distinct from, the cross-lingual and
code-mixing transfer question we study here. Building on this foundation,
four recent 2026-era papers push hidden-state probing further along
different axes. \citet{zhang2025icrprobe} introduce the ICR Score, a metric
that captures each transformer module's \emph{contribution} to the residual
stream update rather than the static hidden state itself, and show this
dynamic signal outperforms static-representation probes with far fewer
parameters. \citet{mrykhin2026pep} propose Prompt Embedding Probes, which
augment a frozen LLM's input with a small number of learnable prompt
embeddings before extracting hidden states, improving in-distribution
detection and remaining effective in pre-generation and cross-model transfer
settings, though the authors note that cross-\emph{dataset} transfer remains
difficult. \citet{aiersilan2026quantized} run a systematic comparison of
linear probes, MLP probes, and several sampling-based detectors (INSIDE
EigenScore, self-consistency, attention entropy) across three 7--8B
open-weight models under 4-bit quantization, finding that a single linear
probe on a mid-network layer (blocks 13--18 of 32 for Llama and Mistral;
19--25 of 28 for Qwen) achieves 0.904--1.000 AUROC and that MLP probes rarely
improve on this by more than 0.01 AUROC - a finding we also test in our
own linear-vs-MLP comparison (Section~\ref{sec:experiments}).
\citet{alvi2026multihaludet} take the most direct step toward multilinguality,
introducing a stacking framework that probes full hidden-state trajectories
across layers and evaluating cross-lingual generalization from English to
French, Bangla, and Amharic, reporting strong transfer across this
typologically diverse set.

Across all four of these 2026 papers, and the earlier foundational work they
build on, one gap is consistent: every evaluation is conducted on
\emph{monolingual} input, whether the language is English, French, Bangla,
Amharic, or another single language per example. None test whether the
truthfulness signal survives \emph{within-sentence code-mixing}, where a
single input alternates between two languages at the word or phrase level.
Cross-lingual transfer (e.g., English-trained probe evaluated on Bangla
text) and code-mixing robustness are related but distinct generalization
challenges: the former asks whether a direction learned in one language's
representational geometry recovers in another language's geometry, while the
latter asks whether that direction remains coherent when both geometries are
interleaved within the same forward pass. Our work is, to our knowledge, the
first to test hidden-state hallucination probes under this second condition.

\paragraph{Code-mixing and Hinglish NLP.}
Code-mixing has long been studied from a theoretical linguistics
perspective; the Matrix Language Frame (MLF) model of
\citet{myers-scotton1993duelling} formalizes intrasentential code-switching
in terms of a dominant \emph{matrix language} that supplies the grammatical
frame into which content words from an \emph{embedded language} are
inserted, a framework we adopt for our synthetic Hinglish construction
(Section~\ref{sec:method}). On the NLP benchmarking side, LinCE
\citep{aguilar2020lince} and GLUECoS \citep{khanuja2020gluecos} established
centralized benchmarks for evaluating code-switched NLP systems on tasks
such as language identification, part-of-speech tagging, named entity
recognition, and sentiment analysis across several code-mixed language
pairs, including Hindi--English. These benchmarks establish that code-mixing
is a first-class, well-studied NLP phenomenon with dedicated evaluation
infrastructure - infrastructure that, to date, has not been connected to
the hidden-state hallucination-probing literature reviewed above. We were
unable to obtain access to naturally-occurring code-mixed data from these or
comparable sources during data collection (Section~\ref{sec:limitations}),
and instead construct synthetic Hinglish; we see validating our findings on
naturally-occurring corpora such as LinCE or GLUECoS as the most direct and
important piece of future work.

\paragraph{Multilingual hallucination and factuality evaluation.}
More broadly, a separate line of work evaluates LLM factuality and
hallucination rates across languages without probing internal
representations, typically via multilingual QA or fact-verification
benchmarks built on resources such as XQuAD \citep{artetxe2020xquad} and
IndicQA \citep{doddapaneni2023indicqa}, both of which we draw on for our
clean-language data (Section~\ref{sec:method}). This literature generally
finds that LLMs are less factually reliable in lower-resource languages than
in English, which is consistent with, and provides context for, our own
secondary finding (RQ2) that all three models we evaluate hallucinate far
more often in Hindi and Hinglish than in English on matched questions.

\section{Method}
\label{sec:method}

\subsection{Data construction}
\label{sec:data}

We construct a 5{,}674-item question-answering benchmark spanning three
language conditions derived from comparable underlying facts: clean Hindi,
clean English, and synthetic Hindi--English code-mixed (Hinglish).

\paragraph{Clean Hindi and English.} We combine XQuAD-hi
\citep{artetxe2020xquad} and IndicQA-hi \citep{doddapaneni2023indicqa} for
clean Hindi context-question-answer triples, and XQuAD-en for a matched
clean-English condition, yielding 3{,}432 clean QA pairs (2{,}242 Hindi,
1{,}190 English). We filter IndicQA's SQuAD-2.0-style unanswerable questions,
which are returned as an empty-string answer rather than an empty answer
list, to ensure every retained item has a genuine gold answer.

\paragraph{Synthetic Hinglish.} We were unable to obtain naturally-occurring
Hindi--English code-mixed QA data during data collection (see
Section~\ref{sec:limitations}); the Hindi--English configurations of LinCE
\citep{aguilar2020lince} were unreachable via their external hosting
infrastructure at the time of this work, and no comparable GLUECoS
\citep{khanuja2020gluecos} mirror was accessible. We therefore construct
synthetic Hinglish following the Matrix Language Frame (MLF) model of
\citet{myers-scotton1993duelling}: Hindi is retained as the \emph{matrix
language} supplying the grammatical frame, while a controlled fraction of
\emph{content} words - nouns, verbs, and other open-class items - are
substituted with their English translations drawn from the MUSE
Hindi--English bilingual dictionary \citep{conneau2018muse}. Function words
(pronouns, postpositions, copulas, conjunctions, and other closed-class
items) are explicitly excluded from substitution, since code-mixing that
alters function words produces text that is linguistically unnatural and
not representative of how Hindi--English bilinguals actually code-mix. We
apply this procedure to the 2{,}242 clean-Hindi items, targeting a 40\%
content-word substitution rate per context and per question; the rate
actually achieved varies by item depending on MUSE dictionary coverage. This
yields 2{,}242 synthetic Hinglish items, for a combined dataset of 5{,}674
rows across all three conditions.

\subsection{Hallucination labeling}
\label{sec:labeling}

For each of three open-weight instruction-tuned LLMs - Qwen2.5-7B-Instruct,
Mistral-7B-Instruct-v0.3, and Llama-3.1-8B-Instruct \citep{grattafiori2024llama3}
- we generate an answer to every item in all three language conditions,
using each model's native chat template with a system prompt instructing
short, direct, context-grounded answers, and greedy decoding. This yields
$5{,}674 \times 3 = 17{,}022$ model responses. We label each response as
\emph{faithful} or \emph{hallucinated} via token-level F1 against the gold
answer \citep{rajpurkar2016squad}, the standard metric used in prior
hidden-state probing work \citep{aiersilan2026quantized}, with faithful
defined as F1 $\geq 0.5$.

We note that early pilot runs using a raw string-concatenation prompt rather
than each model's chat template produced degenerate, document-continuation
style generations for all three instruction-tuned models - an artifact of
prompting an instruct model outside its trained input format rather than a
genuine capability failure. We verified this on a small held-out sample
before the full labeling run and confirmed that chat-template prompting
resolves it; we flag this explicitly, since silently mislabeling such
artifacts as hallucination would inflate hallucination rates in a way
unrelated to our research questions.

\subsection{Hidden-state extraction}
\label{sec:extraction}

For each labeled (context, question, generated-answer) triple, we perform a
single teacher-forced forward pass over the concatenation of the
chat-templated prompt and the model's own generated answer, and extract the
hidden state at every transformer layer (plus the embedding layer) at two
token positions: the \emph{last-prompt-token} position, immediately before
generation begins, and the \emph{last-answer-token} position, at the end of
the generated answer. We test both positions because prior work differs on
which is more informative: pre-generation probing tests whether the model
already encodes a signal that it may be about to hallucinate, while
post-generation probing tests whether the completed answer's representation
encodes its own faithfulness. Hidden states are stored in half precision to
limit storage overhead. This yields, per model, one 29-layer (Qwen2.5-7B) or
33-layer (Mistral-7B, Llama-3.1-8B) hidden-state vector at each of the two
positions for every generated response - 5{,}674 responses for Qwen2.5-7B
and Mistral-7B, and 5{,}544 for Llama-3.1-8B after excluding rows lost to an
occasional out-of-memory condition during single-row extraction
($\sim$2\% of rows; see Section~\ref{sec:limitations}).

\subsection{Probing setup}
\label{sec:probing-setup}

For each model, language condition, token position, and layer, we train a
linear probe - logistic regression with balanced class weighting and
standard feature scaling, since unscaled raw hidden-state features are
poorly conditioned for L-BFGS optimization and converge extremely slowly
without it - to predict the faithful/hallucinated label from the
corresponding hidden-state vector. We report three sets of results:

\begin{itemize}
    \item \textbf{In-distribution detection}: 3-fold stratified
    cross-validated AUROC within each (model, language, position, layer)
    combination, reported as a layer-wise curve per model and position
    (Figure~\ref{fig:layerwise}).
    \item \textbf{Cross-lingual transfer to Hinglish}: a probe is trained on
    all available clean-language data (Hindi or English, separately) at a
    given layer and position, then evaluated on the held-out Hinglish
    condition at the same layer and position. This experiment directly
    answers RQ1.
    \item \textbf{Linear-vs-MLP comparison}: at the best-performing layer
    identified by the in-distribution experiment, we additionally train a
    single-hidden-layer (64-unit) MLP probe under the same cross-validation
    protocol, to test whether a linear decision boundary is sufficient or
    whether nonlinear probing recovers meaningfully more signal, following
    the comparison reported by \citet{aiersilan2026quantized}.
\end{itemize}

Conditions with fewer than 10 examples in the minority class are excluded
from in-distribution evaluation, and conditions with fewer than 5 examples
in either class are excluded from transfer evaluation, since AUROC estimated
on very few minority-class examples is unstable. As we report in
Section~\ref{sec:experiments}, class imbalance is substantial for Hindi and
Hinglish in particular, since all three models hallucinate far more
frequently in these conditions than in English; we return to the
implications of this imbalance for interpreting the resulting AUROC values
in Section~\ref{sec:discussion}.

\section{Experiments and Results}
\label{sec:experiments}

\subsection{In-distribution detection}
\label{sec:results-indist}

\begin{figure}[t]
    \centering
    \includegraphics[width=\linewidth]{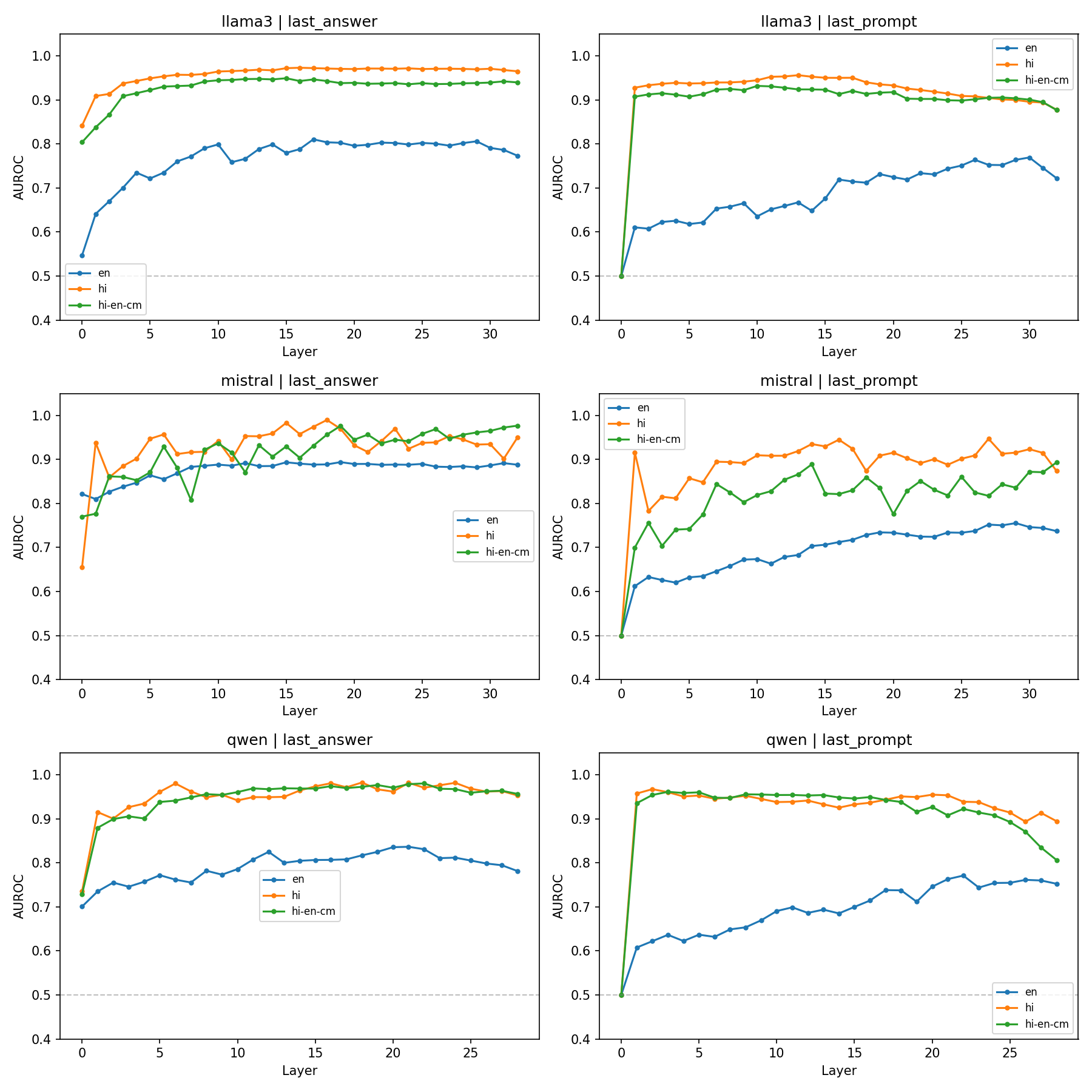}
    \caption{Layer-wise in-distribution AUROC for each model (rows) and
    token position (columns), by language condition. Dashed line marks
    chance level (AUROC = 0.5).}
    \label{fig:layerwise}
\end{figure}

Figure~\ref{fig:layerwise} shows layer-wise AUROC for all three models, both
token positions, and all three language conditions. Two patterns are
consistent across every model and position. First, AUROC rises sharply from
the embedding layer and plateaus within the first few transformer blocks,
consistent with prior findings that truthfulness becomes linearly decodable
early and remains so through most of the network
\citep{aiersilan2026quantized}. Second, and more strikingly for our
purposes, \textbf{Hindi and Hinglish probes substantially outperform English
probes in-distribution across all three models and both positions}: at the
best layer, English AUROC tops out at 0.77--0.84, while Hindi and Hinglish
reach 0.89--0.99 (full best-layer values in Table~\ref{tab:mlp}). We discuss
this pattern, and the class-imbalance caveat relevant to interpreting it, in
Section~\ref{sec:discussion}.

\subsection{Cross-lingual transfer to Hinglish}
\label{sec:results-transfer}

Table~\ref{tab:transfer} reports our central result: for each model and
position, we train a probe on clean Hindi or clean English at the
best-performing layer for that training condition, and evaluate it on
held-out Hinglish. Transfer AUROC ranges from 0.876 (Qwen, English-trained,
prompt position) to 0.991 (Mistral, Hindi-trained, answer position) -
consistently well above chance and, in most conditions, close to what a
probe trained directly on Hinglish achieves in-distribution.

\begin{table}[t]
    \centering
    \small
    \begin{tabular}{llrr}
        \toprule
        Model & Train & Layer & AUROC \\
        \midrule
        \multicolumn{4}{l}{\textit{last-answer-token position}} \\
        Llama-3.1 & en & 19 & 0.924 \\
        Llama-3.1 & hi & 15 & 0.963 \\
        Mistral & en & 15 & 0.939 \\
        Mistral & hi & 19 & \textbf{0.991} \\
        Qwen2.5 & en & 27 & 0.934 \\
        Qwen2.5 & hi & 16 & 0.964 \\
        \midrule
        \multicolumn{4}{l}{\textit{last-prompt-token position}} \\
        Llama-3.1 & en & 12 & 0.898 \\
        Llama-3.1 & hi & 14 & 0.953 \\
        Mistral & en & 14 & 0.917 \\
        Mistral & hi & 2 & 0.939 \\
        Qwen2.5 & en & 15 & 0.876 \\
        Qwen2.5 & hi & 14 & 0.943 \\
        \bottomrule
    \end{tabular}
    \caption{Cross-lingual transfer to Hinglish: probes trained on clean
    Hindi (\texttt{hi}) or clean English (\texttt{en}) at the
    best-performing layer for that training language, evaluated on held-out
    Hinglish. In every one of the six model$\times$position combinations,
    the Hindi-trained probe transfers better than the English-trained
    probe.}
    \label{tab:transfer}
\end{table}

Table~\ref{tab:gap} makes the comparison to in-distribution performance
explicit. The gap between transfer AUROC and in-distribution Hinglish AUROC
(at each model's own best in-distribution layer and position) is under 0.05
AUROC in 10 of 12 conditions, and is \emph{negative} - i.e., transfer
\emph{exceeds} in-distribution performance - in all six Hindi-trained
conditions except two, where it is nonetheless close to zero. The largest
gap we observe is 0.086 AUROC (Qwen2.5, English-trained, prompt position).
Across every one of the six model$\times$position pairs, \textbf{the
Hindi-trained probe transfers more reliably to Hinglish than the
English-trained probe}, with a mean gap of $-0.017$ for Hindi-trained probes
versus $+0.043$ for English-trained probes.

\begin{table}[t]
    \centering
    \small
    \begin{tabular}{llrr}
        \toprule
        Model & Position & Train & Gap \\
        \midrule
        Llama-3.1 & answer & en & $+0.025$ \\
        Llama-3.1 & answer & hi & $-0.014$ \\
        Llama-3.1 & prompt & en & $+0.034$ \\
        Llama-3.1 & prompt & hi & $-0.021$ \\
        Mistral & answer & en & $+0.037$ \\
        Mistral & answer & hi & $-0.014$ \\
        Mistral & prompt & en & $-0.024$ \\
        Mistral & prompt & hi & $-0.046$ \\
        Qwen2.5 & answer & en & $+0.046$ \\
        Qwen2.5 & answer & hi & $+0.017$ \\
        Qwen2.5 & prompt & en & $+0.086$ \\
        Qwen2.5 & prompt & hi & $+0.019$ \\
        \bottomrule
    \end{tabular}
    \caption{Gap = transfer AUROC $-$ in-distribution Hinglish AUROC (at the
    model's own best in-distribution layer/position). Negative values mean
    transfer \emph{outperforms} in-distribution detection. Hindi-trained
    probes (gap column, \texttt{hi} rows) are closer to zero or negative in
    5 of 6 cases; English-trained probes are positive (i.e., a real
    degradation) in 5 of 6 cases.}
    \label{tab:gap}
\end{table}

Taken together, these two tables directly answer RQ1: \textbf{the
truthfulness signal survives code-mixing well}. We find no evidence of the
signal collapsing or becoming unreliable under Hinglish input for any of the
three model families or two probing positions we test.

\subsection{Linear vs.\ MLP probes}
\label{sec:results-mlp}

Table~\ref{tab:mlp} compares linear and MLP probes at each model's
best-performing layer per language and position. Consistent with
\citet{aiersilan2026quantized}, MLP probes rarely improve substantially on
the linear baseline, and in several low-resource conditions (particularly
Hinglish and Hindi at the prompt position for Mistral) the MLP
\emph{underperforms} the linear probe by a wide margin, most likely due to
overfitting on the smaller, highly imbalanced minority class discussed in
Section~\ref{sec:probing-setup}. We take this as further support for using
linear probes as the primary, most reliable method, with MLP probes serving
only as a secondary sanity check.

\begin{table}[t]
    \centering
    \small
    \begin{tabular}{llrr}
        \toprule
        Model / Lang & Position & Linear & MLP \\
        \midrule
        Llama-3.1 / en & prompt & 0.769 & 0.770 \\
        Llama-3.1 / hi & prompt & 0.956 & 0.961 \\
        Llama-3.1 / cm & prompt & 0.932 & 0.934 \\
        Mistral / en & prompt & 0.755 & 0.785 \\
        Mistral / hi & prompt & 0.947 & 0.890 \\
        Mistral / cm & prompt & 0.893 & \textbf{0.699} \\
        Qwen2.5 / en & prompt & 0.771 & 0.779 \\
        Qwen2.5 / hi & prompt & 0.968 & 0.965 \\
        Qwen2.5 / cm & prompt & 0.962 & 0.956 \\
        \bottomrule
    \end{tabular}
    \caption{Linear vs.\ MLP probe AUROC at each condition's best
    in-distribution layer (prompt position shown; answer position follows
    the same pattern and is reported in full in
    Table~\ref{tab:mlp-full}, Appendix~\ref{sec:appendix-mlp}). ``cm''
    denotes Hinglish. The MLP's sharp drop for Mistral/Hinglish
    illustrates overfitting under class imbalance rather than a genuine
    linear-probe ceiling.}
    \label{tab:mlp}
\end{table}

\subsection{Faithful-rate comparison across languages}
\label{sec:results-faithful}

\begin{figure}[t]
    \centering
    \includegraphics[width=\linewidth]{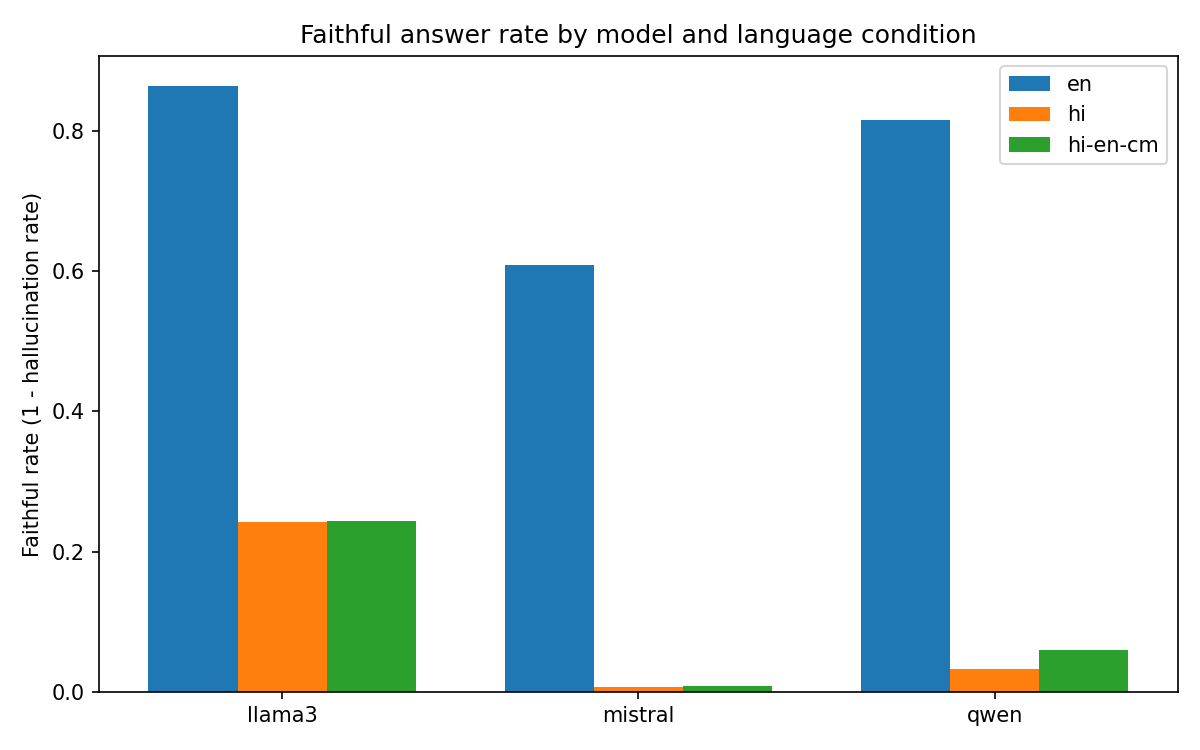}
    \caption{Faithful (non-hallucinated) answer rate by model and language
    condition. All three models hallucinate far more often in Hindi and
    Hinglish than in English.}
    \label{fig:faithful}
\end{figure}

As an independent, practically motivating finding addressing RQ2,
Figure~\ref{fig:faithful} shows the fraction of faithful (non-hallucinated)
answers per model and language. All three models are dramatically less
faithful in Hindi and Hinglish than in English: Llama-3.1 drops from 86.4\%
faithful in English to 24.2\% in Hindi and 24.4\% in Hinglish; Mistral drops
from 60.9\% to under 1\% in both Hindi and Hinglish; and Qwen2.5 drops from
81.5\% to 3.3\% (Hindi) and 5.9\% (Hinglish). Hindi and Hinglish faithful
rates are close to each other for every model, suggesting that
code-mixing itself does not substantially change how often these models
hallucinate relative to clean Hindi - the large faithfulness gap is
primarily a Hindi-vs-English gap, not a code-mixing-specific effect.

\section{Discussion}
\label{sec:discussion}

\paragraph{Why does the signal survive code-mixing?} We did not expect, a
priori, that transfer to Hinglish would be this robust. One plausible
explanation is that the truthfulness direction identified by these probes is
encoded relatively early and is tied more to task-level semantic content
(whether the stated fact matches the context) than to surface tokenization
or script. Since our synthetic Hinglish retains the Hindi grammatical frame
and only swaps content words, the underlying semantic content the model must
track for truthfulness is largely preserved even as surface form changes -
which may explain why the signal is more robust to code-mixing than one
might initially expect from a purely surface-level generalization argument.

\paragraph{Why do Hindi-trained probes transfer better than English-trained
probes?} Across all six model$\times$position pairs, Hindi-trained probes
transfer to Hinglish more reliably than English-trained probes
(Table~\ref{tab:transfer}, Table~\ref{tab:gap}). We hypothesize this
reflects a shared-script and shared-grammatical-frame effect: our synthetic
Hinglish retains Hindi as its matrix language, so a probe trained on clean
Hindi representations is being asked to generalize to text that shares
grammatical structure and (for the majority, non-substituted portion of each
sentence) surface form with its training distribution, whereas an
English-trained probe must generalize across a much larger representational
shift. We present this as a hypothesis motivated by our construction
procedure rather than a claim we independently verify; testing it directly
- for instance by varying the matrix language or the substitution rate -
is a natural direction for follow-up work.

\paragraph{Practical implication.} For practitioners building hallucination
detectors intended for real-world, code-mixed traffic, our central finding
is reassuring: a detector validated only on clean-language data is not
obviously broken when it encounters Hinglish input, at least for the
in-sentence code-mixing pattern and language pair we study. This does not
mean such a detector should be deployed without further validation (see
Limitations below), but it does mean the code-mixing gap identified in
Section~\ref{sec:related} is, on the evidence here, not the silent failure
mode one might have feared.

\paragraph{Interpreting the Hindi/Hinglish AUROC advantage.} The
in-distribution results in Section~\ref{sec:results-indist} show Hindi and
Hinglish probes reaching substantially higher AUROC than English probes.
Part of this advantage is likely genuine - the faithful-rate results in
Section~\ref{sec:results-faithful} show these models are far less capable in
Hindi and Hinglish, and a larger, more separable gap between confident-and-
correct and confident-and-wrong internal states is a plausible consequence
of operating further outside a model's core competence. However, we caution
that severe class imbalance (as low as 17 faithful examples out of 2{,}242
for Mistral/Hindi) makes AUROC estimates in these conditions noisier and
potentially more optimistic than the equivalent estimates for the more
balanced English condition, even with class-weighted training and
stratified cross-validation. We do not believe this fully explains the
pattern - it is consistent across three independently trained models and
two independent token positions - but we flag it as a factor future work
should control for, for instance by evaluating on class-balanced subsamples
or reporting confidence intervals over resampled folds.

\section*{Limitations}
\label{sec:limitations}

\paragraph{Synthetic rather than natural code-mixing.} Our Hinglish data is
synthetically constructed via matrix-language-frame word substitution
(Section~\ref{sec:data}), not drawn from naturally-occurring code-mixed
text. We attempted to obtain natural Hindi--English code-mixed QA data from
LinCE \citep{aguilar2020lince} and GLUECoS \citep{khanuja2020gluecos} during
data collection but were unable to access either at the time of this work.
Naturally-occurring code-mixing exhibits patterns - inconsistent
switch-points, borrowed discourse markers, transliteration variation,
speaker-specific mixing styles - that our controlled substitution
procedure does not capture. Our central finding (the signal survives
code-mixing) should therefore be read as evidence about a controlled,
linguistically-motivated approximation of Hinglish, and validating it on
naturally-occurring code-mixed corpora is, in our view, the single most
important piece of follow-up work.

\paragraph{Single language pair.} We study only Hindi--English code-mixing.
Whether these findings generalize to other code-mixed language pairs, with
different scripts, typological distances, or sociolinguistic mixing
patterns, is untested.

\paragraph{Class imbalance.} As discussed in Section~\ref{sec:discussion},
Hindi and Hinglish faithful/hallucinated labels are severely imbalanced for
two of our three models (Mistral and Qwen2.5), which affects the
reliability of AUROC estimates in those conditions despite our use of
class-weighted training, stratified cross-validation, and minimum
class-count thresholds for excluding underpowered conditions.

\paragraph{Data loss in extraction.} Hidden-state extraction for
Llama-3.1-8B completed for 5{,}544 of 5{,}674 rows ($\sim$97.7\%), with the
remainder excluded due to an occasional out-of-memory condition during
single-row forward-pass extraction. We do not expect this small, effectively
random subset of missing rows to bias our results, but note it for
completeness.

\paragraph{Token-F1 as a labeling proxy.} We label hallucination via
token-F1 against a single gold answer, following standard practice in prior
hidden-state probing work \citep{aiersilan2026quantized}. This is a proxy
for genuine faithfulness, not a human-verified ground truth: it can
penalize correct answers phrased differently from the reference, and cannot
detect hallucinated content in answers that happen to contain the correct
key term. We expect this labeling noise to affect all three language
conditions similarly rather than to selectively bias the Hindi-vs-Hinglish
comparison central to our claims, but it remains a limitation of the
evaluation protocol we inherit from prior work rather than one we resolve.

\section*{Ethics Statement}

This work aims to improve hallucination detection for a large,
currently underserved population of Hindi--English bilingual LLM users, and
we believe better detection tools for this population have clear positive
potential. At the same time, we emphasize that hidden-state probes of the
kind we study are not perfect classifiers and should not be treated as a
sole or final safety mechanism in any deployed system; a probe reporting
high confidence that an answer is faithful does not guarantee correctness,
and the risks of over-trusting such a signal, particularly in
high-stakes applications, should be weighed carefully. Our synthetic
Hinglish dataset is derived entirely from existing, publicly available QA
benchmarks (XQuAD, IndicQA) and a publicly available bilingual dictionary
(MUSE); it does not involve human subjects or personally identifiable data,
and we release it to support, not substitute for, further validation on
naturally-occurring code-mixed data.

\section{Conclusion}

We present the first evaluation of hidden-state hallucination probing under
within-sentence code-mixed input, addressing a gap left open by every prior
study of this probing paradigm we are aware of. Across three open-weight
7--8B LLMs, two token positions, and a controlled synthetic Hindi--English
Hinglish benchmark, we find that the truthfulness signal these probes rely
on survives code-mixing well: cross-lingual transfer AUROC to Hinglish
ranges from 0.876 to 0.991, with a gap to in-distribution performance under
0.05 AUROC in most conditions, and probes trained on clean Hindi transfer
more reliably than probes trained on clean English across every model and
position we test. As an independent finding, all three models we evaluate
hallucinate substantially more often in Hindi and Hinglish than in English
on matched questions, underscoring the practical importance of reliable
hallucination detection for this language setting. We release our code and
synthetic Hinglish QA benchmark, and identify validation on
naturally-occurring code-mixed corpora as the clearest direction for future
work.

\bibliography{references}

\appendix
\section{Appendix}

\subsection{Prompt template}
We use the following system prompt, combined with each model's native chat
template, for all generation: \textit{``You are a helpful assistant that
answers questions based only on the given context. Give a short, direct
answer only - no explanation, no extra text, no additional questions.''}
The user turn contains the context, question, and an explicit
\texttt{Answer:} cue. Decoding is greedy (temperature-independent) for all
models and all conditions.

\subsection{Compute and environment}
Data generation and hidden-state extraction were run on an NVIDIA RTX A6000
(48GB) and, for one model, on Kaggle's dual T4 (16GB$\times$2) instances
using 4-bit quantization. Probe training and analysis are CPU-only and
complete in minutes given the extracted hidden states. All three LLMs were
run in their publicly released instruction-tuned form with no additional
fine-tuning.

\subsection{Full per-layer results}
Complete per-layer AUROC values for all (model, language, position) triples,
underlying Figure~\ref{fig:layerwise}, along with the full transfer results
for both token positions underlying Table~\ref{tab:transfer} and
Table~\ref{tab:gap}, are released with our code and data.

\subsection{Full linear-vs-MLP comparison}
\label{sec:appendix-mlp}
Table~\ref{tab:mlp-full} reports the complete linear-vs-MLP comparison for
all three models, all three language conditions, and both token positions
(Table~\ref{tab:mlp} in the main text shows only the prompt-position subset).
The answer-position results follow the same overall pattern: MLP probes
rarely improve meaningfully on the linear baseline, and the largest
MLP underperformance again occurs for Mistral on Hinglish, though the drop
is smaller at the answer position than at the prompt position.

\begin{table}[h]
    \centering
    \small
    \begin{tabular}{llrr}
        \toprule
        Model / Lang & Position & Linear & MLP \\
        \midrule
        Llama-3.1 / en & prompt & 0.769 & 0.770 \\
        Llama-3.1 / hi & prompt & 0.956 & 0.961 \\
        Llama-3.1 / cm & prompt & 0.932 & 0.934 \\
        Llama-3.1 / en & answer & 0.811 & 0.818 \\
        Llama-3.1 / hi & answer & 0.973 & 0.976 \\
        Llama-3.1 / cm & answer & 0.949 & 0.955 \\
        \midrule
        Mistral / en & prompt & 0.755 & 0.785 \\
        Mistral / hi & prompt & 0.947 & 0.890 \\
        Mistral / cm & prompt & 0.893 & 0.699 \\
        Mistral / en & answer & 0.894 & 0.903 \\
        Mistral / hi & answer & 0.990 & 0.982 \\
        Mistral / cm & answer & 0.977 & 0.966 \\
        \midrule
        Qwen2.5 / en & prompt & 0.771 & 0.779 \\
        Qwen2.5 / hi & prompt & 0.968 & 0.965 \\
        Qwen2.5 / cm & prompt & 0.962 & 0.956 \\
        Qwen2.5 / en & answer & 0.837 & 0.850 \\
        Qwen2.5 / hi & answer & 0.983 & 0.954 \\
        Qwen2.5 / cm & answer & 0.981 & 0.977 \\
        \bottomrule
    \end{tabular}
    \caption{Full linear-vs-MLP probe AUROC comparison at each condition's
    best in-distribution layer, both token positions. ``cm'' denotes
    Hinglish.}
    \label{tab:mlp-full}
\end{table}

\end{document}